\documentclass[]{spie}  

\usepackage{amsfonts}
\usepackage{amsmath}
\usepackage{multirow} 
 
\usepackage{amsmath,amsfonts,amssymb}
\usepackage{graphicx}
\usepackage[colorlinks=true, allcolors=blue]{hyperref}

\title{Unsupervised Domain Adaptation for Segmentation with Black-box Source Model}

\author[a]{Xiaofeng Liu}
\author[a,b]{Chaehwa Yoo}
\author[a]{Fangxu Xing}
\author[c]{C.-C. Jay Kuo}
\author[a]{Georges El Fakhri}
\author[a,b]{Je-Won Kang}
\author[a]{Jonghye Woo}

\affil[a]{Gordon Center for Medical Imaging, Massachusetts
General Hospital and Harvard Medical School, Boston, MA, USA}
\affil[b]{Dept. of Electronic and Electrical Engineering and Graduate Program in Smart Factory, Ewha Womans University, Seoul, South Korea}
\affil[c]{Dept. of Electrical and Computer Engineering, University of Southern California, Los Angeles, CA, USA}

\begin{document} 
\maketitle

\begin{abstract}
Unsupervised domain adaptation (UDA) has been widely used to transfer knowledge from a labeled source domain to an unlabeled target domain to counter the difficulty of labeling in a new domain. The training of conventional solutions usually relies on the existence of both source and target domain data. However, privacy of the large-scale and well-labeled data in the source domain and trained model parameters can become the major concern of cross center/domain collaborations. In this work, to address this, we propose a practical solution to UDA for segmentation with a black-box segmentation model trained in the source domain only, rather than original source data or a white-box source model. Specifically, we resort to a knowledge distillation scheme with exponential mixup decay (EMD) to gradually learn target-specific representations. In addition, unsupervised entropy minimization is further applied to regularization of the target domain confidence. We evaluated our framework on the BraTS 2018 database, achieving performance on par with white-box source model adaptation approaches.

\end{abstract}


\section{Introduction}

Accurate tumor segmentation is an important prerequisite for early diagnosis and treatment planning, which has been substantially improved with advances in deep learning approaches. The performance of a pre-trained model, however, can be severely degraded, when a source distribution differs from a target distribution. For example, tumors with different grades are likely to yield different distributions, which hampers the application of a single segmentation model in segmenting tumors of varying degrees of severity and growth~\cite{liu2021generative}. In addition, it is costly to annotate high-quality labeled data in the new target domain. Unsupervised domain adaptation (UDA) has been proposed to address the problem of domain shift~\cite{liu2021subtype}. In UDA, a segmentation model is typically trained using both source and target data, but only the source data are labeled. 

Although UDA offers a promising solution to the problem of domain shift, due to the concerns over leakage of sensitive information contained in the patient data, it is often challenging to access source data, when deploying the developed model in the target domain~\cite{liu2021adapting}. To address this, Liu et al. proposed a source-free or source-relaxed UDA (i.e., white-box domain adaptation) for segmentation~\cite{liu2021adapting}. In that work, an off-the-shelf segmentation model was adapted to a target domain via a pre-trained model in a source domain, by transferring its batch normalization statistics. Recently, a deep inversion technique \cite{yin2020dreaming}, however, has shown that original training data can be recovered from knowledge used during white-box domain adaptation, which may leak confidential information \cite{zhang2021unsupervised}. Accordingly, this work aims to address the limitation, by developing a black-box domain adaptation approach that further restricts the use of knowledge from a source segmentation model. 

To our knowledge, this is the first attempt at achieving UDA for a deep segmentation network using black-box domain adaptation. The black-box setting provides a more effective way to protect privacy, compared with white-box domain adaptation approaches \cite{liu2021adapting} or conventional UDA approaches~\cite{zou2019confidence}. Recently, Zhang et al. \cite{zhang2021unsupervised} proposed to use the black-box UDA for classification, with the class-wise noise rate estimation and category-wise sampling. However, that work is not applicable to the segmentation task to carry out pixel-wise classification. In previous works, segmentation based on the black-box domain adaptation was largely ignored, although it could be used in a more realistic and challenging scenario.

\section{Related work}

Unsupervised domain adaptation \cite{liu2021subtype,liu2021energy,he2020classification,he2020image2audio} has been an important technology to address the difficulties such as domain shift and costly labeling in the new domains. Conventional approaches utilized both source and target domain data for training \cite{liu2021adversarial,liu2021generative,liu2021recursively,liu2021dual,liu2021unified}. Recently, source-free UDA \cite{wang2020fully,liang2020we,bateson2020source} has been proposed to only use the pre-trained model rather than co-training the network with source and target domain data. A recent study \cite{liu2021Off-the-Shelf} explored the shared or domain-specific batch-normalization statistics to achieve domain alignment. We note that domain generalization \cite{liu2021domain}, a closely related but different task, assumes that there are no target domain data in its training. 

In this work, we target to tackle a more restrictive scenario, in which we only use a black-box pre-trained model, and do not rely on the network parameters. As stated above, a deep inversion technique \cite{yin2020dreaming}, has shown that original training data can be recovered from knowledge during white-box domain adaptation \cite{zhang2021unsupervised}. A contemporary work \cite{zhang2021unsupervised} proposed to adapt a black-box classification model to a target domain without both source domain data and the model structure. That work proposed iterative learning with noisy labels, which regarded the black-box predictions as noisy labels. To the best of our knowledge, black-box UDA for segmentation has not been investigated. The semantic segmentation \cite{liu2020importance,liu2020severity,liu2020wasserstein,liu2021segmentation,wang2021automated,liu2020reinforced} makes pixel-wise classification and provides more explainable results for the subsequent decision making process.

\section{METHOD}

\begin{figure}[t]
\begin{center}
\includegraphics[width=1\linewidth]{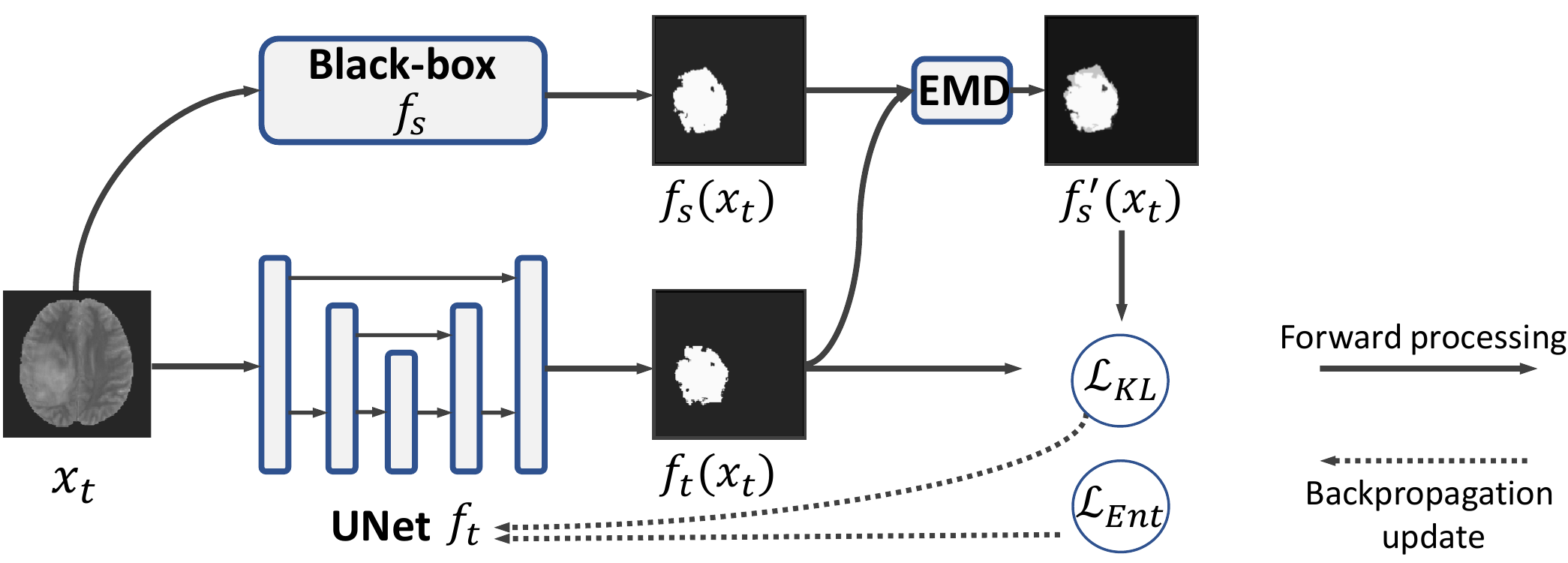} 
\end{center} 
\caption{Illustration of our black-box UDA framework using knowledge distillation with EMD pseudo label and unsupervised entropy minimization.}  
\label{ccc}\end{figure}

In the black-box UDA segmentation task, we have a network $f_s$ trained with the labeled source domain sample $\{x_s,y_s\}$, where $f_s$ is fixed and accessed only through a nontransparent API at the adaptation stage. At the adaptation stage, we only have access to the black-box $f_s$ and the unlabeled target domain sample $\{x_t\}$ to train a target domain network $f_t$ to achieve a good segmentation performance in the target domain. In this work, we propose a practical solution to the black-box UDA for segmentation with a noise-aware knowledge distillation scheme using pseudo label with exponential mixup decay (EMD). The framework is shown in Fig. \ref{ccc}.

Following the knowledge distillation \cite{yin2020dreaming}, the well-trained source model can act as a teacher to provide its pixel-wise softmax histogram prediction of each image. The target domain model is trained to imitate the source model. The consistency of their prediction can be enforced with the Kullback–Leibler (KL) divergence between their pixel-wise softmax histogram distributions. However, due to the domain shift, the prediction of $f_s$ can be noisy. Accordingly, we resort to a self-training scheme \cite{liu2021generative} to construct the pseudo label for target domain training. Specifically, to achieve the gradual translation to the target domain, we mix up the source and target domain predictions, i.e., $f_s(x_t)$ and $f_t(x_t)$, and adjust their ratio for the pseudo label $y_t'$ with EMD: 
\begin{align}
{y_t'}_n = \lambda f_s(x_t)_n + (1-\lambda) f_t(x_t)_n,
\end{align} 
where $n$ indexes the pixel, and $f_s(x_t)_n$ and $f_t(x_t)_n$ are the histogram distributions of the softmax output of the $n$-th pixel of the predictions $f_s(x_t)$ and $f_t(x_t)$, respectively. $\lambda=\lambda^0\text{exp}(-I)$ is a target adaptation momentum parameter with an exponential decay w.r.t. the iteration $I$. $\lambda^0$ is the initial weight of $f_s(x_t)$, which is empirically set to 1. The loss knowledge distillation with the EMD pseudo label can be formulated as: 
\begin{align}
    \mathcal{L}_{KL}= \frac{1}{H_0\times W_0} \sum_n^{H_0\times W_0} \mathcal{D}_{KL}(f_t(x_t)_n||{y_t'}_n),
\end{align} 
where $H_0$ and $W_0$ are the height and width of the image. Therefore, the weight of $\lambda$ can be smoothly decreased along with the training, and $f_t$ gradually represents the target data.

In addition, an unsupervised training scheme, e.g., entropy minimization \cite{grandvalet2005semi}, can be added on top of our framework. For implementation, the entropy for pixel segmentation can be formulated as an averaged entropy of the pixel-wise softmax prediction, given by 
\begin{align}
    \mathcal{L}_{Ent}=
    \frac{1}{ H_0\times W_0} \sum_n^{H_0\times W_0}\{{f_t(x_t)_n \text{log} f_t(x_t)_n}\}.
\end{align} 
Minimizing $\mathcal{L}_{Ent}$ leads to the output $f_t(x_t)_n$ close to a one-hot distribution, i.e., confident prediction. 

In summary, our training objective can be formulated as
\begin{align}
    \mathcal{L}={L}_{KL}+\alpha\mathcal{L}_{Ent},
\end{align}where $\alpha$ is used to balance between the knowledge distillation with the EMD pseudo label and entropy minimization. We note that a trivial solution to entropy minimization is that all unlabeled target data could have the same one-hot encoding \cite{grandvalet2005semi}. Thus, in order to stabilize the training, we linearly change the hyper-parameter $\alpha$ from 5 to 0.

\section{RESULTS}

We evaluated our approach on the BraTS2018 database, comprising a total of 210 high-grade glioma (HGG) subjects and a total of 75 low-grade glioma (LGG) subjects. \cite{menze2014multimodal}. Each subject has T1-weighted (T1), T1-contrast enhanced (T1ce), T2-weighted (T2), and T2 Fluid Attenuated Inversion Recovery (FLAIR) Magnetic Resonance Imaging (MRI) volumes with voxel-wise labels for the enhancing tumor (EnhT), the peritumoral edema (ED), and the necrotic and non-enhancing tumor core (CoreT). We denote the sum of EnhT, ED, and CoreT as the whole tumor. More information about
the database can be found in \cite{menze2014multimodal}. Training was performed on four NVIDIA TITAN Xp GPUs with the PyTorch deep learning toolbox~\cite{paszke2017automatic}, which took about 5 hours.

Following the previous white-box source free UDA \cite{liu2021adapting} and UDA with source data \cite{shanis2019intramodality}, we used HGG subjects as the source domain and the LGG subjects as the target domain, which have different size and position distributions \cite{shanis2019intramodality}. We trained $f_s$ using the prior work \cite{liu2021adapting}, and did not have access to its network parameters at the adaptation stage. For simplicity, we used the same 2D U-Net backbone for $f_s$ and $f_t$ as in \cite{shanis2019intramodality}. Specifically, we used U-Net \cite{ronneberger2015u} as our segmentation network with 15 layers, batch normalization, and dropout. The network was trained using Adam as an
optimizer with $\beta_1= 0.9$ and $\beta_2= 0.99$.

The slices of four modalities were concatenated as a 4-channel input with the spatial size of 128$\times$128. The training of adaptation used the LGG training set. The evaluation was implemented in the LGG testing set. For evaluation, we adopted the two metrics, including Dice similarity coefficient (DSC) and Hausdorff distance (HD) metrics \cite{zou2020unsupervised}.
 
\begin{figure}[t]
\begin{center}\vspace{+5pt}
\includegraphics[width=1\linewidth]{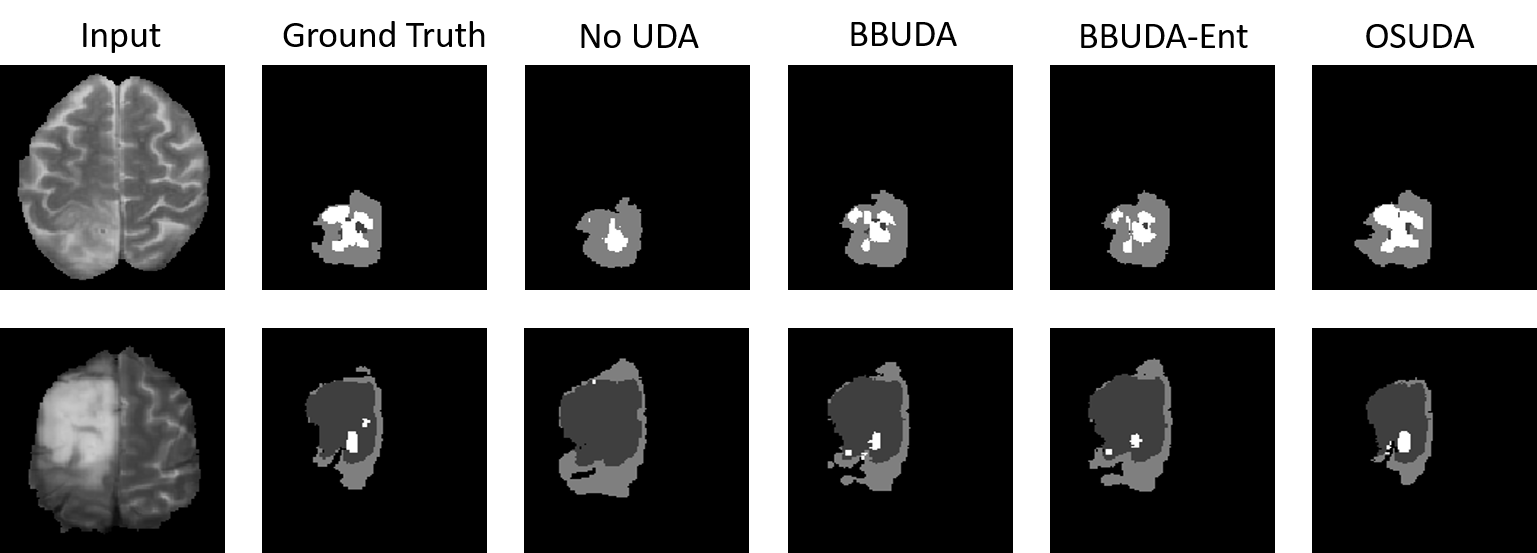} 
\end{center}
\caption{Comparison with the other UDA methods, and an ablation study of entropy minimization for HGG to LGG UDA. We use white, dark gray, and gray to indicate the CoreT, EnhT and ED, respectively. OSUDA \cite{liu2021adapting} with the white-box source model for UDA training is regarded as an ``upper bound."}  
\label{exp1}
\end{figure} 

We used BBUDA to denote our black-box UDA framework, and BBUDA-Ent indicates the ablation study without entropy minimization. An illustration of the segmentation results is shown in Fig. \ref{exp1}. We can see that the predictions of our proposed BBUDA are better than the no adaptation model. The better performance of BBUDA over BBUDA-Ent demonstrates the effectiveness of our entropy minimization.

The quantitative evaluation results are shown in Table~\ref{tab1}. Our proposed BBUDA achieved the state-of-the-art performance for the black-box source-free UDA segmentation, approaching performance of the white-box OSUDA~\cite{bateson2020source,liu2021adapting} with the source model parameters, which can be considered an ``upper-bound." We note that the label ratio consistency assumption in CRUDA \cite{bateson2020source} does not hold for this HGG to LGG transfer task, which thus led to an inferior performance.

\begin{table}[t!]\vspace{+10pt}
\caption{Comparison of HGG to LGG black-box UDA with the four-channel input for our four-class segmentation, i.e., whole tumor, enhanced tumor, core tumor, and background. OSUDA \cite{liu2021adapting} with the white-box source model for UDA training is regarded as an ``upper bound."} 
\label{tab1}\vspace{+5pt}
\centering
\resizebox{0.8\linewidth}{!}{
\begin{tabular}{l|c|ccc|ccc}
    \hline
    \multirow{2}*{Method}&Source& \multicolumn{3}{c|}{Dice Score [\%] $\uparrow$}& \multicolumn{3}{c}{Hausdorff Distance [mm] $\downarrow$} \\ \cline{3-8}
    &model& WholeT & EnhT & CoreT   & WholeT & EnhT & CoreT   \\ \hline \hline
     Source only~\cite{liu2021adapting} & no UDA &79.29& 30.09& 44.11   &38.7 &46.1 &40.2  \\\hline  
    BBUDA & black-box &82.21 & 31.33& 46.64   &28.6 &26.4 &28.1  \\ 
    BBUDA-Ent & black-box &81.84 & 31.26& 45.75   &29.4 &27.5 &29.0  \\ \hline      

    CRUDA~\cite{bateson2020source} & white-box & 79.85 &31.05& 43.92  &31.7 &29.5 &30.2  \\ 
   
    {OSUDA}~\cite{liu2021adapting} & white-box &{83.62} & {32.15}& {46.88}   &{27.2} &{23.4} &{26.3}  \\ \hline
\end{tabular}
}
\end{table}

In addition to the HGG to LGG setting, we also proposed to adapt from LGG to HGG. Similar to the HGG to LGG task, there was also the domain gap w.r.t. tumor types and the label proportion of each class. The results are shown in Table \ref{tab2}. Our proposed BBUDA achieved superior segmentation performance consistently.



\section{CONCLUSION}

In this work, we tackled the problem of black-box UDA for segmentation under a realistic and meaningful scenario, and proposed a practical and efficient knowledge distillation scheme with EMD pseudo label. Specifically, we were able to smoothly transfer the segmentation in the source domain to the target domain with EMD to construct the pseudo label. In addition, unsupervised entropy minimization was added on top of our model to further boost the performance. Experimental results, carried out with the HGG to LGG adaptation task, showed that our proposed BBUDA outperformed the source model, by a large margin, and importantly, the DSC and HD metrics were comparable to the white-box UDA approaches. In this work, we only investigated the adaptation between HGG and LGG data, while the model can be general for any segmentation UDA tasks. In addition, more advanced knowledge distillation and unsupervised learning methods can be added on to further boost the performance, which is subject to our future work.

\begin{table}[t!]\vspace{+5pt}
\caption{Comparison of LGG to HGG black-box UDA with the four-channel input for our four-class segmentation, i.e., whole tumor, enhanced tumor, core tumor, and background. OSUDA \cite{liu2021adapting} with the white-box source model for UDA training is regarded as an ``upper bound."} 
\label{tab2}\vspace{+5pt}
\centering
\resizebox{0.8\linewidth}{!}{
\begin{tabular}{l|c|ccc|ccc}
    \hline
    \multirow{2}*{Method}&Source& \multicolumn{3}{c|}{Dice Score [\%] $\uparrow$}& \multicolumn{3}{c}{Hausdorff Distance [mm] $\downarrow$} \\ \cline{3-8}
    &model& WholeT & EnhT & CoreT   & WholeT & EnhT & CoreT   \\ \hline \hline
    
     Source only~\cite{liu2021adapting} & no UDA &81.45& 34.36& 40.30   &36.7 &41.6 &37.2  \\\hline  
    BBUDA & black-box &85.47 & 39.56& 45.18   &26.7 &33.8 &29.6  \\ 
    BBUDA-Ent & black-box &84.92 & 38.64& 44.73   &27.1&34.6 &31.3 \\ \hline      

    CRUDA~\cite{bateson2020source} & white-box & 87.62 &40.17& 49.65 &23.9&22.7 &23.9  \\ 
   
    {OSUDA}~\cite{liu2021adapting} & white-box &{89.75} & {44.21}& {50.34}   &{22.2} &{19.3} &{21.6}  \\ \hline
\end{tabular}
}
\end{table}

\acknowledgments 
This work is partially supported by NIH R01DC018511 and P41EB022544.

\bibliography{report} 
\bibliographystyle{spiebib} 

\end{document}